\documentclass[10pt,twocolumn,letterpaper]{article}

\PassOptionsToPackage{table,xcdraw}{xcolor}

\usepackage[pagenumbers]{cvpr}

\usepackage{microtype}

\renewcommand{\paragraph}[1]{\vspace{.5em}\noindent\textbf{#1.}}

\definecolor{cvprblue}{rgb}{0.21,0.49,0.74}
\usepackage[breaklinks,colorlinks,allcolors=cvprblue]{hyperref}
\usepackage{bookmark}

\usepackage{pifont}
\usepackage{wrapfig}
\usepackage{siunitx}
\usepackage{multirow}
\usepackage[normalem]{ulem}
\usepackage{stfloats}
\newcommand{\cmark}{\ding{51}}

\definecolor{realgrayhighlight}{gray}{0.95}
\definecolor{grayhighlight}{HTML}{F6F6F6}
\definecolor{bestcell}{HTML}{D0E0EE}
\definecolor{secondcell}{HTML}{EAEDF1}
\DeclareRobustCommand{\high}[1]{\cellcolor{ForestGreen!20}\textbf{#1}}
\DeclareRobustCommand{\med}[1]{\cellcolor{LimeGreen!20}\underline{#1}}
\DeclareRobustCommand{\highcap}[1]{{\setlength{\fboxsep}{1.5pt}\colorbox{ForestGreen!20}{\textbf{#1}}}}
\DeclareRobustCommand{\medcap}[1]{{\setlength{\fboxsep}{1.5pt}\colorbox{LimeGreen!20}{#1}}}

\title{ReCoSplat: Online Feed-Forward Gaussian Splatting via Render-and-Compare}

\author{
Freeman Cheng$^{1}$ \quad Botao Ye$^{2}$ \quad Xueting Li$^{3}$ \\
Junqi You$^{4}$ \quad Fangneng Zhan$^{5}$ \quad Ming-Hsuan Yang$^{1}$ \\[0.65em]
{\small $^{1}$University of California, Merced \quad $^{2}$ETH Z\"urich \quad $^{3}$NVIDIA} \\
{\small $^{4}$Shanghai Jiao Tong University \quad $^{5}$HKUST} \\[0.55em]
{\small
\href{https://freemancheng.com/ReCoSplat/}{\raisebox{-0.12em}{\includegraphics[height=0.82em]{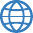}}\hspace{0.28em}Project Page}
\hspace{0.75em}{\color{black!35}\textbullet}\hspace{0.75em}
\href{https://github.com/peelstnac/ReCoSplat}{\raisebox{-0.12em}{\includegraphics[height=0.82em]{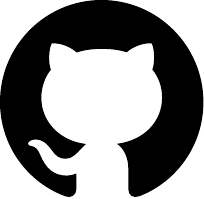}}\hspace{0.28em}Code}}
}

\hypersetup{
    pdftitle={ReCoSplat: Online Feed-Forward Gaussian Splatting via Render-and-Compare},
    pdfauthor={Freeman Cheng, Botao Ye, Xueting Li, Junqi You, Fangneng Zhan, Ming-Hsuan Yang},
    pdfsubject={Online feed-forward 3D scene reconstruction with Gaussian splatting},
    pdfkeywords={online novel view synthesis, feed-forward 3D reconstruction, Gaussian splatting, camera pose estimation},
    pdfdisplaydoctitle=true,
    bookmarksopen=true,
    bookmarksopenlevel=2,
    bookmarksnumbered=true
}

\begin{document}
\maketitle

\pdfbookmark[1]{Abstract}{abstract}
\begin{abstract}
\upshape
Online novel view synthesis requires a model to reconstruct a scene causally from a stream of observations while keeping it renderable at every moment.
We present \textbf{ReCoSplat}, an online feed-forward Gaussian Splatting model supporting both posed and unposed inputs, with or without camera intrinsics.
While assembling local Gaussians with camera poses scales better than canonical-space prediction, stable training requires ground-truth poses, creating a distribution mismatch when predicted poses are used at inference.
To address this, we introduce a Render-and-Compare (ReCo) module.
ReCo renders the accumulated scene from the viewpoint of the incoming observation, comparing the render with the observation to produce a stable conditioning signal that helps bridge the mismatch.
To support long sequences, we propose a hybrid KV-cache compression strategy combining early-layer truncation with chunk-level selective retention, reducing the KV cache size by over 90\% for 100 or more frames.
ReCoSplat achieves state-of-the-art performance among online methods while processing 256-view streams at an average input throughput of 45.1 FPS, with an end-of-stream throughput of 41.1 FPS on an RTX 6000 Ada GPU.
Code and pretrained models are released at the \href{https://freemancheng.com/ReCoSplat/}{project page}.
\end{abstract}

\suppressfloats[t]

\begin{figure}[t]
    \centering
    \includegraphics[width=\linewidth]{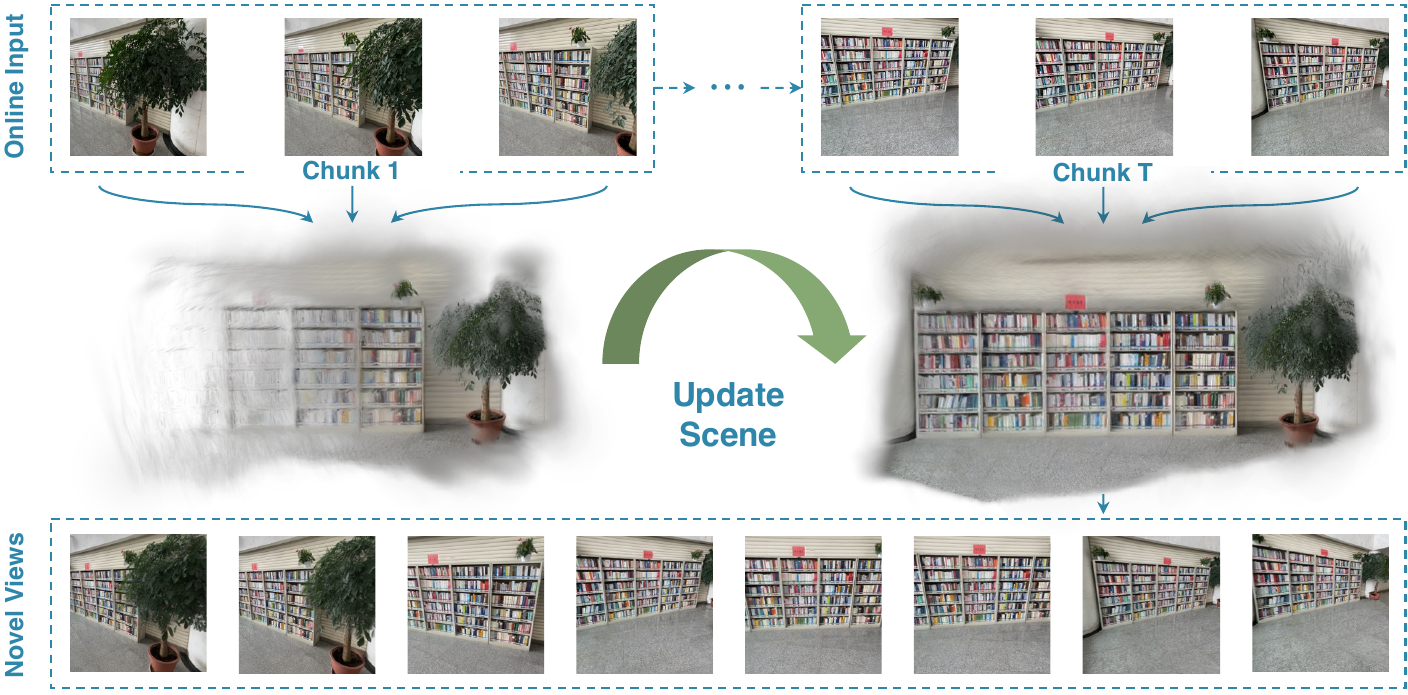}
    \caption{\textbf{Online reconstruction setting.} ReCoSplat receives images sequentially and causally updates a persistent 3D Gaussian scene without access to future observations. The accumulated scene remains renderable after every update.}
    \label{fig:teaser}
\end{figure}

\section{Introduction}
\label{sec:intro}

Novel view synthesis aims to render photorealistic images of a scene from unseen viewpoints. 
Among scene representations, 3D Gaussians have been widely adopted due to their high-quality real-time rendering. 
However, optimization-based Gaussian Splatting \cite{kerbl3Dgaussians} requires lengthy per-scene training, limiting real-time applications. 
Feed-forward Gaussian Splatting \cite{charatan23pixelsplat, szymanowicz24splatter, gslrm2024, ye2025no, jiang2025anysplat, xu2024depthsplat, ye2025yonosplatneedmodelfeedforward} predicts 3D Gaussians directly from input images, bypassing per-scene optimization. 
Yet these approaches target offline reconstruction, where all images are available upfront. 
The \emph{online} setting remains far less explored despite its importance for embodied AI, AR/VR, and video generation \cite{chen2026contextforcingconsistentautoregressive, wu2025video}. 
An online model must therefore provide two capabilities: \textbf{causal processing}, incorporating new observations without access to future frames, and \textbf{anytime rendering}, keeping the scene representation renderable at every moment of the input stream.

To this end, we present ReCoSplat, an online feed-forward Gaussian Splatting method for scene reconstruction from hundreds of images in posed or unposed settings, with or without camera intrinsics.

Recent work demonstrates that predicting Gaussians in local camera space and mapping them to world coordinates via camera poses, which we term \textbf{assembly poses}, is more scalable than predicting Gaussians in a canonical space \cite{ye2025no, ye2025yonosplatneedmodelfeedforward}. 
Yet a critical design choice remains: whether to use ground-truth or predicted assembly poses during training, a decision that divides existing online methods.
StreamGS \cite{StreamGS} and SaLon3R \cite{SaLon3R} use predicted poses, but this leads to training instability \cite{ye2025yonosplatneedmodelfeedforward} by coupling Gaussian prediction with pose estimation.
Using ground-truth poses decouples the tasks, as in LongSplat \cite{huang2025longsplatonlinegeneralizable3d}, but assumes camera poses are provided at inference.
A natural alternative is to train with ground-truth poses and switch to predicted poses at inference, but this creates a \textbf{pose distribution mismatch} that can misalign the assembled Gaussians.
Although offline models such as YoNoSplat \cite{ye2025yonosplatneedmodelfeedforward} address this mismatch through training curricula, we find that mixing ground-truth and predicted assembly poses is ineffective in the online setting, where pose estimation is more challenging.

To address this mismatch, \textbf{ReCo}Splat introduces a \textbf{Re}nder-and-\textbf{Co}mpare (ReCo) module inspired by the Analysis-by-Synthesis framework \cite{YUILLE2006301}. 
Following this principle, ReCo renders the accumulated scene at the incoming observation’s assembly pose and compares the resulting render with the observation to condition its local Gaussian prediction.
By exposing discrepancies between the render and the incoming observation, ReCo helps the Gaussian predictor generalize from ground-truth assembly poses during training to predicted poses at inference.
Surprisingly, ReCo yields even larger gains when ground-truth poses are provided, as local Gaussian prediction errors are no longer entangled with pose-estimation errors.
To enrich the comparison signal, we augment each Gaussian with learned feature channels that are rendered alongside RGB.

To sustain anytime rendering over hundreds of frames, we address the KV-cache memory bottleneck in transformer backbones such as \(\pi^3\) \cite{pi3}.
Our hybrid compression strategy combines early-layer truncation with chunk-level selective retention, reducing the KV-cache size by over 90\% for streams of 100 or more frames while preserving reconstruction quality.
ReCoSplat processes 256-view streams at an average input throughput of 45.1 FPS, with an end-of-stream throughput of 41.1 FPS on an RTX 6000 Ada GPU.

The contributions of this work are:
\begin{itemize}
    \item \textbf{Render-and-Compare Module.}
    We condition local Gaussian prediction on a rendered comparison between the accumulated scene and each incoming observation, helping bridge the mismatch between ground-truth assembly poses during training and predicted poses at inference.

    \item \textbf{Efficient Long-Sequence Reconstruction.}
    Our hybrid KV-cache compression strategy reduces the cache size by over 90\% for 100 or more input frames while preserving quality, leading to an end-of-stream throughput of 41.1 FPS for 256-view streams on an RTX 6000 Ada.

    \item \textbf{Causal, Anytime-Renderable Reconstruction.}
    ReCoSplat supports posed or unposed inputs, with or without camera intrinsics, while maintaining an explicit Gaussian scene after every update, and achieves state-of-the-art performance among online methods across these settings.
\end{itemize}

\section{Related Work}
\label{sec:related_work}

\paragraph{Feed-Forward Multi-View Stereo}
Feed-forward Multi-View Stereo (MVS) methods predict dense geometry directly from RGB images.
Early methods such as MVSNet~\cite{yao2018mvsnet, yao2019recurrentmvsnet} construct cost volumes but require known camera parameters.
DUSt3R~\cite{Wang_2024_CVPR} removes this constraint by predicting 3D pointmaps from image pairs in the first-view camera space.
Later methods extend to multi-view settings using alternating attention~\cite{wang2025vggt, keetha2026mapanything, pi3}, with~$\pi^3$ showing that canonical first-view coordinate spaces introduce harmful inductive bias. 
Depth Anything 3~\cite{depthanything3} instead uses a standard transformer architecture over alternating attention to better exploit pretrained weights.
Several works further extend MVS to online inputs through sliding windows~\cite{li2025wint3rwindowbasedstreamingreconstruction}, spatial pointer memory~\cite{wu2025pointr}, accumulated or pruned KV~caches~\cite{streamVGGT, yuan2026infinitevggt}, recurrent state memory~\cite{cut3r}, and test-time training~\cite{chen2025ttt3r}. 
While these pointmap methods excel at geometry reconstruction, they cannot be used for novel view synthesis without further processing, motivating our online feed-forward Gaussian Splatting approach.

\paragraph{Feed-Forward Gaussian Splatting}
Feed-forward Gaussian Splatting predicts 3D Gaussian primitives from images without per-scene optimization~\cite{zhang2025advancesfeedforward3dreconstruction}.
Early methods establish the pixel-aligned formulation, mapping each input pixel to a Gaussian in a single forward pass~\cite{charatan23pixelsplat, szymanowicz24splatter, gslrm2024}.
Subsequent methods leverage pretrained models to improve reconstruction quality~\cite{ye2025no, xu2024depthsplat}, scale to more input views~\cite{jiang2025anysplat, ye2025yonosplatneedmodelfeedforward}, and relax input requirements~\cite{ye2025no, jiang2025anysplat, ye2025yonosplatneedmodelfeedforward}.
Other methods redesign the model architecture itself to support longer inputs.
For example, Long-LRM~\cite{ziwen2025llrm} adopts Mamba~\cite{mamba2} blocks and token merging, while ZPressor~\cite{wang2025zpressor} condenses the information from dense inputs into a small subset of views.

\paragraph{Online Feed-Forward Gaussian Splatting}
Recent work extends feed-forward Gaussian Splatting to the online setting. 
Most of these methods follow a predict-and-merge design, predicting Gaussians from each incoming frame and merging them into an accumulated scene that remains renderable as new frames arrive.
To stabilize prediction from unposed streams, StreamGS~\cite{StreamGS} matches features across frames with adaptive density control, while OF$^3$GS~\cite{chen2026of3gs} predicts per-pixel offsets that let Gaussians deviate from their unprojection rays.
To manage the accumulated scene, SaLon3R~\cite{SaLon3R} compresses Gaussians into saliency-aware anchors refined by a point transformer, and LongSplat~\cite{huang2025longsplatonlinegeneralizable3d} encodes the reconstruction history into a Gaussian-Image Representation (GIR).
LongSplat is the most closely related to our design, yet differs in several aspects.
First, LongSplat requires ground-truth poses for Gaussian assembly.
Second, it uses the GIR to encode historical context, whereas our Render-and-Compare module improves generalization from ground-truth to predicted assembly poses.
Third, using the GIR within our ReCo module degrades performance.

tttLRM~\cite{wang2026tttlrm} takes a different route, using test-time training to compress incoming posed images into a fixed-size fast-weight state.
It obtains a renderable scene by decoding per-pixel Gaussians from this state.
Because each state update leaves previously decoded Gaussians outdated, maintaining an anytime-renderable scene requires re-decoding the entire view history after every update, a cost that quickly outpaces the incoming stream.
ReCoSplat instead decodes Gaussians only for the incoming images and merges them into an explicit accumulated scene, which remains renderable after every update.
Unlike tttLRM, ReCoSplat also supports unposed inputs.

\section{Method}
\label{sec:method}

\begin{figure*}[t]
    \centering
    \includegraphics[width=0.93\linewidth]{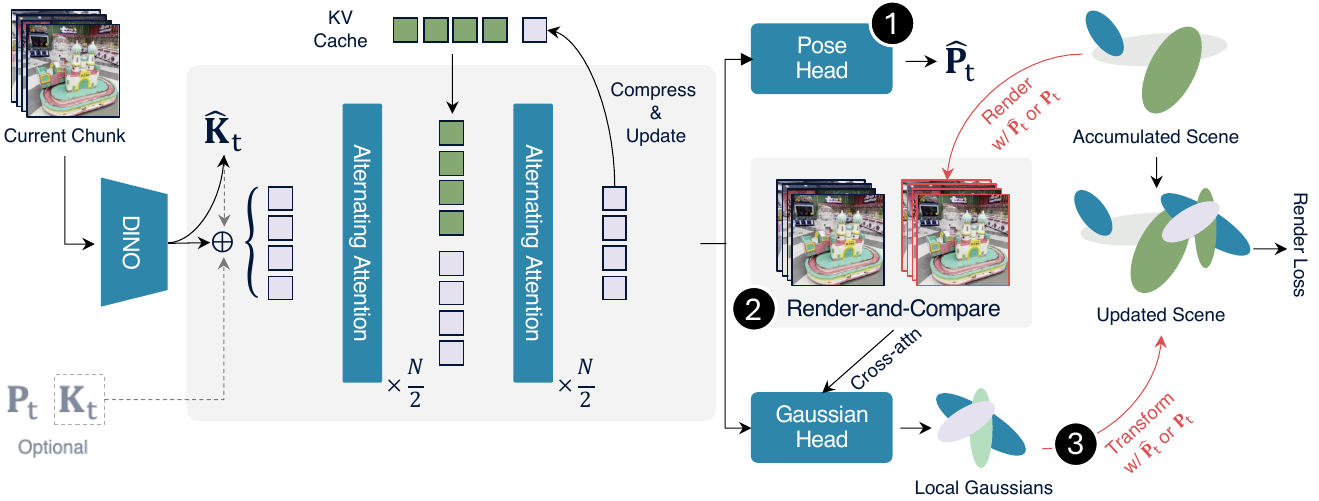}
    \caption{\textbf{ReCoSplat overview.}
A DINO encoder and an alternating-attention transformer extract multi-view features from the incoming chunk, using a compressed KV cache to attend to past chunks (Sec.~\ref{sec:kv-cache-compression}).
(1)~The pose head predicts camera poses.
Assembly poses are ground-truth poses when available and the predictions otherwise (Eq.~\ref{eq:assembly-pose}).
(2)~ReCo renders the accumulated scene $S_{t-1}$ at the assembly poses and compares the renders with the observations to condition local Gaussian prediction.
(3)~The same assembly poses transform the local Gaussians into world coordinates for merging into the updated scene $S_t$.}
    \label{fig:pipeline}
\end{figure*}

\subsection{Online Feed-Forward Gaussian Splatting}
\label{sec:online-formulation}

ReCoSplat (Figure~\ref{fig:pipeline}) reconstructs a scene from an image stream processed in non-overlapping chunks of $n\in\{4,\ldots,8\}$ views.
At timestep $t$, its inputs are the incoming image chunk $\mathbf I_t=(I_t^1,\ldots,I_t^n)$, optional camera parameters (intrinsics $\mathbf K_t$ and extrinsics $\mathbf P_t$), the KV cache $M_{t-1}$, and the accumulated Gaussian scene $S_{t-1}$.
The model predicts local Gaussians $\mathbf G_t=(G_t^1,\ldots,G_t^n)$, camera parameters $\hat{\mathbf K}_t$ and $\hat{\mathbf P}_t$, and updates the KV cache $M_{t-1}\rightarrow M_t$:
\begin{equation}
    (\mathbf G_t,\hat{\mathbf K}_t,\hat{\mathbf P}_t,M_t)
    =f_\theta(\mathbf I_t,\mathbf K_t,\mathbf P_t,M_{t-1},S_{t-1}).
    \label{eq:online-update}
\end{equation}
Note that the incoming chunk is processed with access only to prior chunks (causal processing), and that since $S_t$ is an explicit set of 3D Gaussians, novel views can be rendered after every update (anytime rendering).

As each $G_t^k$ is defined in the local camera space of $I_t^k$, it must be mapped into the world coordinate system with an \textbf{assembly pose} $A_t^k$.
Formally, let $\mathcal T_A(G)$ apply the camera-to-world transformation $A$ to $G$.
The accumulated Gaussian scene is updated as
\begin{equation}
    S_t
    =S_{t-1}
    \cup\mathcal T_{A_t^1}(G_t^1)
    \cup\cdots
    \cup\mathcal T_{A_t^n}(G_t^n).
\end{equation}
During training, ground-truth poses are used as assembly poses, as this teacher-forcing strategy is necessary for stable optimization (Section \ref{sec:ablation}).
At inference, however, assembly poses can be either ground-truth or predicted, depending on whether camera extrinsics are available as input.

\subsection{Render-and-Compare Module}
\label{sec:render-and-compare}

As discussed in Section~\ref{sec:intro}, training with ground-truth assembly poses is necessary but creates a pose distribution mismatch when predicted poses are used at inference (Sections~\ref{sec:results},~\ref{sec:ablation}).
Curricula that mix ground-truth and predicted poses~\cite{ye2025yonosplatneedmodelfeedforward} do not close this gap as predicted assembly poses provide weak supervision in the online setting (Section~\ref{sec:ablation}).
To address the mismatch, we introduce ReCo, which conditions local Gaussian prediction on a rendered comparison between the accumulated scene and the incoming observation.

For each image $I_t^k$ in the incoming chunk, ReCo renders the accumulated scene $S_{t-1}$ at its assembly pose $A_t^k$:
\begin{equation}
    \hat R_t^k=\operatorname{Render}(S_{t-1};A_t^k,\widetilde K_t^k).
    \label{eq:reco-update}
\end{equation}
We use ground-truth intrinsics for $\widetilde K_t^k$ when available and the prediction $\hat K_t^k$ otherwise.
ReCo encodes the rendered--observed pair as $Z_t^k$:
\begin{equation}
    (\hat R_t^k,I_t^k)
    \xrightarrow{\operatorname{Patchify}}
    Z_t^k.
\end{equation}
Then, given multi-view backbone features $F_t^k$ (Section~\ref{sec:autoregressive-backbone}), a Gaussian head predicts $G_t^k$ from $F_t^k$ by conditioning on $Z_t^k$ via cross-attention:
\begin{equation}
    G_t^k=\operatorname{GaussianHead}(F_t^k,Z_t^k).
    \label{eq:gaussian-prediction}
\end{equation}

Because the assembly pose $A_t^k$ is used both to render $\hat R_t^k$ and to map $G_t^k$ into world coordinates, any error in $A_t^k$ is reflected in the rendered--observed pair.
Consequently, assembly-pose errors are directly visible to the Gaussian head, enabling appropriate adjustments to its prediction and improving the model's generalization to predicted poses at inference (Table~\ref{tab:dl3dv-1}).

The rendering $\hat R_t^k\in\mathbb R^{H\times W\times12}$, where $H$ and $W$ are the image height and width, contains RGB and nine learned feature channels.
For the first chunk, where no accumulated scene exists, a learned embedding is used as the render.
$\operatorname{Patchify}$ consists of three convolutional layers with strided downsampling.
In practice, $\operatorname{GaussianHead}$ comprises three separate heads that respectively decode Gaussian positions, attributes, and learned features.
The position and attribute heads are five-layer transformers followed by linear projections, and the feature head is a DPT~\cite{Ranftl_2021_ICCV}.

\subsection{Causal Backbone and Extrinsics Prediction}
\label{sec:autoregressive-backbone}

ReCoSplat builds on the multi-view backbone of YoNoSplat~\cite{ye2025yonosplatneedmodelfeedforward} by replacing its global attention with chunkwise causal attention.
For each input chunk $\mathbf I_t$, the backbone $B_\theta$ extracts corresponding multi-view features $\mathbf F_t$ and updates the KV cache:
$
    (\mathbf F_t,M_t)
    =B_\theta(\mathbf I_t,M_{t-1};\mathbf K_t),
    \label{eq:backbone-update}
$
where $M_{t-1}$ and $M_t$ are the incoming and updated KV caches, respectively, and the provided intrinsics $\mathbf K_t$ are optional.

\paragraph{Image and Intrinsics Embedding}
A ViT encoder~\cite{dosovitskiy2021an} initialized with DINOv2~\cite{oquab2024dinov} maps each image $I_t^k$ to a camera token $c_t^k$ and patch tokens $E_t^k$.
A two-layer MLP predicts the focal lengths $\hat K_t^k$ from $c_t^k$.
The intrinsics, either provided as $K_t^k$ or predicted as $\hat K_t^k$, define a pixel-wise intrinsic embedding that is added to $E_t^k$ before decoding.

\paragraph{Alternating-Attention Decoder}
The decoder, consisting of 36 frame/global alternating attention layers~\cite{pi3}, takes the intrinsics-embedded patch tokens $\mathbf E_t=(E_t^1,\ldots,E_t^n)$ and the KV cache $M_{t-1}$ as input.
Frame attention processes each view independently, while global attention exchanges information across views in the current chunk and with $M_{t-1}$.
The decoder produces $\mathbf F_t=(F_t^1,\ldots,F_t^n)$, where each $F_t^k$ corresponds to an image $I_t^k$ in the input chunk.
After processing a chunk, we update the KV cache $M_{t - 1}\rightarrow M_t$ according to our KV-cache compression strategy (Section~\ref{sec:kv-cache-compression}).

\paragraph{Extrinsics Prediction}
A five-layer frame-wise transformer followed by a two-layer MLP maps each multi-view feature $F_t^k$ to a corresponding camera extrinsics prediction $\hat P_t^k$.

\subsection{Efficient Long-Sequence Reconstruction}
\label{sec:kv-cache-compression}

Scaling ReCoSplat to hundreds of images is constrained by the memory demands of its transformer backbone \cite{pi3}.
Naively accumulating a KV cache across all global attention layers causes VRAM usage to grow rapidly with sequence length, making long-sequence inference impractical on consumer hardware.
We address this bottleneck with two complementary KV-cache compression strategies.

\paragraph{Early Layer Truncation}
Prior analysis~\cite{sun2025avggtrethinkingglobalattention} shows that early global attention layers extract local features rather than establishing multi-view correspondences.
We therefore omit the KV cache for the first 10 of the 18 global attention layers, restricting attention in these layers to the current chunk.

\paragraph{Selective Context Retention}
For the remaining 8 global attention layers, discarding all historical tokens severely degrades reconstruction quality.
We instead retain KV tokens for every image in the first chunk, which establishes the initial scene context, and only the final image of each subsequent chunk of size $n\in\{4,\ldots,8\}$.
To ensure the network effectively utilizes this compressed cache, we assign a distinct, trainable register token to each retained view.

\paragraph{Memory Footprint}
Without compression, all 18 global attention layers cache the tokens of every view.
With compression, only the final eight such layers maintain a cache, each retaining tokens corresponding to $8+\left\lceil\frac{N-8}{n}\right\rceil$ of the $N$ views (all eight views from the first chunk and one view from each subsequent chunk).
For $N=256$ and $n=8$, each caching layer only retains tokens associated with 39 of 256 views, reducing the overall KV cache size by 93\%.
Figure~\ref{fig:memory_scaling} shows the measured VRAM savings.

\subsection{Training and Inference}
\label{sec:training_and_inference}

\paragraph{Stages}
Training proceeds in three stages.
In \textbf{Stage 1}, we initialize from YoNoSplat's~\cite{ye2025yonosplatneedmodelfeedforward} DL3DV-10K~\cite{ling2024dl3dv} checkpoint and train for 150k steps at a learning rate of $5\times10^{-6}$, keeping the encoder frozen.
We sample 2--64 context views per iteration, linearly increasing the maximum from 8 to 64 between steps 1.5k and 75k.
The chunk size is fixed at 8, and KV cache compression is disabled.
In \textbf{Stage 2}, we train for an additional 50k steps at a reduced learning rate of $5\times10^{-7}$ and introduce variable chunk sizes after the first chunk.
Each chunk size is sampled uniformly between a scheduled minimum and 8, with the minimum linearly annealed from 8 to 4 over the first 25k steps.
The first chunk remains fixed at 8 views to establish the initial scene context.
In \textbf{Stage 3}, we train for a final 50k steps at the same learning rate and enable the full KV cache compression strategy described in Section~\ref{sec:kv-cache-compression}.

\paragraph{Losses}
We supervise renderings of the accumulated scene against eight held-out target images using pixel-wise MSE and LPIPS~\cite{zhang2018perceptual}.
Camera estimation is supervised by an $\ell_2$ loss on predicted focal lengths and a pairwise relative-pose loss over the predicted extrinsics, while an $\ell_1$ penalty on Gaussian opacities encourages sparsity.
We set $\lambda_{\mathrm{MSE}}=1.0$, $\lambda_{\mathrm{LPIPS}}=0.05$, $\lambda_{\mathrm{intrinsic}}=0.5$, $\lambda_{\mathrm{extrinsic}}=0.1$, and $\lambda_{\mathrm{opacity}}=0.01$, and prune Gaussians with opacity below $0.005$.
Complete loss formulations are provided in the appendix.

\begin{table*}[t]
\centering
\caption{Novel view synthesis on DL3DV with 32, 64, 128, and 256 input views sampled from the first 180, 240, 300, and (up to) 360 frames of each scene, respectively. The \highcap{best} and \medcap{second best} metrics are marked, excluding \textcolor{gray}{offline} methods.}
\label{tab:dl3dv-1}
\scriptsize
\renewcommand{\arraystretch}{0.95}
\setlength{\tabcolsep}{2pt}
    \begin{tabular}{@{} lcc
        p{0.3em}
        *{3}{c} p{0.3em}
        *{3}{c} p{0.3em}
        *{3}{c} p{0.3em}
        *{3}{c} @{}}
        \toprule
        \multirow{2}{*}{Method} & \multirow{2}{*}{$p$} & \multirow{2}{*}{$k$} & & \multicolumn{3}{c}{32v} & & \multicolumn{3}{c}{64v} & & \multicolumn{3}{c}{128v} & & \multicolumn{3}{c}{256v} \\
        & & & & {PSNR $\uparrow$} & {SSIM $\uparrow$} & {LPIPS $\downarrow$} & & {PSNR $\uparrow$} & {SSIM $\uparrow$} & {LPIPS $\downarrow$} & & {PSNR $\uparrow$} & {SSIM $\uparrow$} & {LPIPS $\downarrow$} & & {PSNR $\uparrow$} & {SSIM $\uparrow$} & {LPIPS $\downarrow$} \\
        \midrule
        
        \textcolor{gray}{YoNoSplat} & & & & \textcolor{gray}{22.368} & \textcolor{gray}{0.736} & \textcolor{gray}{0.180} & & \textcolor{gray}{22.253} & \textcolor{gray}{0.732} & \textcolor{gray}{0.183} & & \textcolor{gray}{21.827} & \textcolor{gray}{0.720} & \textcolor{gray}{0.194} & & \textcolor{gray}{20.749} & \textcolor{gray}{0.677} & \textcolor{gray}{0.226} \\
        \textcolor{gray}{AnySplat} & & & & \textcolor{gray}{19.944} & \textcolor{gray}{0.617} & \textcolor{gray}{0.240} & & \textcolor{gray}{19.864} & \textcolor{gray}{0.609} & \textcolor{gray}{0.249} & & \textcolor{gray}{19.566} & \textcolor{gray}{0.596} & \textcolor{gray}{0.264} & & \textcolor{gray}{N/A$^{\dagger}$} & \textcolor{gray}{N/A$^{\dagger}$} & \textcolor{gray}{N/A$^{\dagger}$} \\
        OF$^3$GS  & & & & {17.002} & {0.451} & {0.479} & & {16.519} & {0.432} & {0.505} & & {16.095} & {0.414} & {0.530} & & {15.609} & {0.399} & {0.550} \\
        KV Cache  & & & & {21.705} & \med{0.703} & \med{0.202} & & {21.302} & {0.680} & {0.223} & & {20.784} & {0.655} & {0.247} & & {19.819} & {0.606} & {0.292} \\
        GIR       & & & & \med{21.752} & \med{0.703} & {0.203} & & \med{21.382} & \med{0.682} & \med{0.222} & & \med{20.933} & \med{0.660} & \med{0.245} & & \med{19.900} & \med{0.608} & \med{0.291} \\
        \rowcolor{grayhighlight}
        Ours      & & & & \high{22.097} & \high{0.716} & \high{0.194} & & \high{21.774} & \high{0.696} & \high{0.213} & & \high{21.284} & \high{0.672} & \high{0.235} & & \high{20.220} & \high{0.617} & \high{0.281} \\
        
        \midrule
        
        \textcolor{gray}{YoNoSplat} & & \textcolor{gray}{\cmark} & & \textcolor{gray}{22.575} & \textcolor{gray}{0.748} & \textcolor{gray}{0.177} & & \textcolor{gray}{22.514} & \textcolor{gray}{0.747} & \textcolor{gray}{0.179} & & \textcolor{gray}{22.053} & \textcolor{gray}{0.735} & \textcolor{gray}{0.189} & & \textcolor{gray}{20.953} & \textcolor{gray}{0.692} & \textcolor{gray}{0.221} \\
        S3PO-GS       & & \cmark & & {15.598} & {0.399} & {0.541} & & {16.231} & {0.431} & {0.539} & & {16.862} & {0.451} & {0.529} & & {16.809} & {0.446} & {0.533} \\
        OF$^3$GS  & & \cmark & & {17.062} & {0.456} & {0.476} & & {16.558} & {0.437} & {0.503} & & {16.134} & {0.418} & {0.527} & & {15.677} & {0.403} & {0.547} \\
        KV Cache  & & \cmark & & {21.932} & \med{0.717} & \med{0.198} & & {21.544} & \med{0.697} & \med{0.218} & & {20.993} & {0.670} & {0.242} & & {20.007} & \med{0.622} & \med{0.286} \\
        GIR   & & \cmark & & \med{21.947} & {0.716} & {0.199} & & \med{21.600} & \med{0.697} & \med{0.218} & & \med{21.107} & \med{0.672} & \med{0.241} & & \med{20.085} & \med{0.622} & \med{0.286} \\
        \rowcolor{grayhighlight}
        Ours      & & \cmark & & \high{22.417} & \high{0.734} & \high{0.188} & & \high{22.068} & \high{0.713} & \high{0.206} & & \high{21.576} & \high{0.690} & \high{0.227} & & \high{20.430} & \high{0.633} & \high{0.275} \\
        
        \midrule
        
        \textcolor{gray}{YoNoSplat} & \textcolor{gray}{\cmark} & \textcolor{gray}{\cmark} & & \textcolor{gray}{22.998} & \textcolor{gray}{0.781} & \textcolor{gray}{0.162} & & \textcolor{gray}{22.978} & \textcolor{gray}{0.784} & \textcolor{gray}{0.161} & & \textcolor{gray}{22.597} & \textcolor{gray}{0.779} & \textcolor{gray}{0.167} & & \textcolor{gray}{21.549} & \textcolor{gray}{0.749} & \textcolor{gray}{0.190} \\
        \textcolor{gray}{ZPressor} & \textcolor{gray}{\cmark} & \textcolor{gray}{\cmark} & & \textcolor{gray}{19.293} & \textcolor{gray}{0.624} & \textcolor{gray}{0.269} & & \textcolor{gray}{18.568} & \textcolor{gray}{0.586} & \textcolor{gray}{0.297} & & \textcolor{gray}{17.992} & \textcolor{gray}{0.557} & \textcolor{gray}{0.321} & & \textcolor{gray}{17.539} & \textcolor{gray}{0.532} & \textcolor{gray}{0.342} \\
        KV Cache  & \cmark & \cmark & & {22.392} & {0.758} & \med{0.177} & & {22.210} & {0.754} & \med{0.184} & & {21.731} & \med{0.739} & \med{0.199} & & {20.694} & \med{0.699} & \med{0.235} \\
        GIR       & \cmark & \cmark & & \med{22.478} & \med{0.760} & \med{0.177} & & \med{22.264} & \med{0.755} & {0.185} & & \med{21.779} & \med{0.739} & {0.200} & & \med{20.743} & \med{0.699} & \med{0.235} \\
        \rowcolor{grayhighlight}
        Ours      & \cmark & \cmark & & \high{23.084} & \high{0.780} & \high{0.164} & & \high{23.086} & \high{0.782} & \high{0.167} & & \high{22.852} & \high{0.777} & \high{0.176} & & \high{22.003} & \high{0.751} & \high{0.202} \\
        
        \bottomrule
    \end{tabular}
\end{table*}

\paragraph{Scale Alignment}
Ground-truth poses serve as assembly poses during training and posed inference.
Because ground-truth extrinsics need not be metric, their translation scale may differ from that of the model's internal predictions.
Before Gaussian assembly, we therefore rescale the ground-truth translations to match the model's predicted trajectory by aligning the scales of their first chunk.
Let $\mathbf P_1=\{P_1^k\}_{k=1}^{n}$ and $\hat{\mathbf P}_1=\{\hat P_1^k\}_{k=1}^{n}$ denote the ground-truth and predicted poses of the first chunk.
For any set of poses $\mathbf Q$, define $\operatorname{Scale}(\mathbf Q)$ as the maximum distance between its camera centers.
For view $k$ in chunk $t$, write its ground-truth pose as $P_t^k=[r_t^k,t_t^k]$ and denote its predicted pose by $\hat P_t^k$.
The assembly pose is then
\begin{equation}
    A_t^k=
    \begin{cases}
        \left[r_t^k,
        \dfrac{\operatorname{Scale}(\hat{\mathbf P}_1)}
              {\operatorname{Scale}(\mathbf P_1)}t_t^k\right],
        & \text{posed inference},\\
        \hat P_t^k, & \text{unposed inference}.
    \end{cases}
    \label{eq:assembly-pose}
\end{equation}
It is necessary to use the first chunk for scale alignment. 
Unlike offline methods that can normalize the full pose trajectory \cite{ye2025yonosplatneedmodelfeedforward}, an online model has no access to future poses and must commit to a scale at the very first chunk.

\section{Experiments}
\label{sec:experiments}

\begin{figure*}[t]
    \centering
    \includegraphics[width=\linewidth]{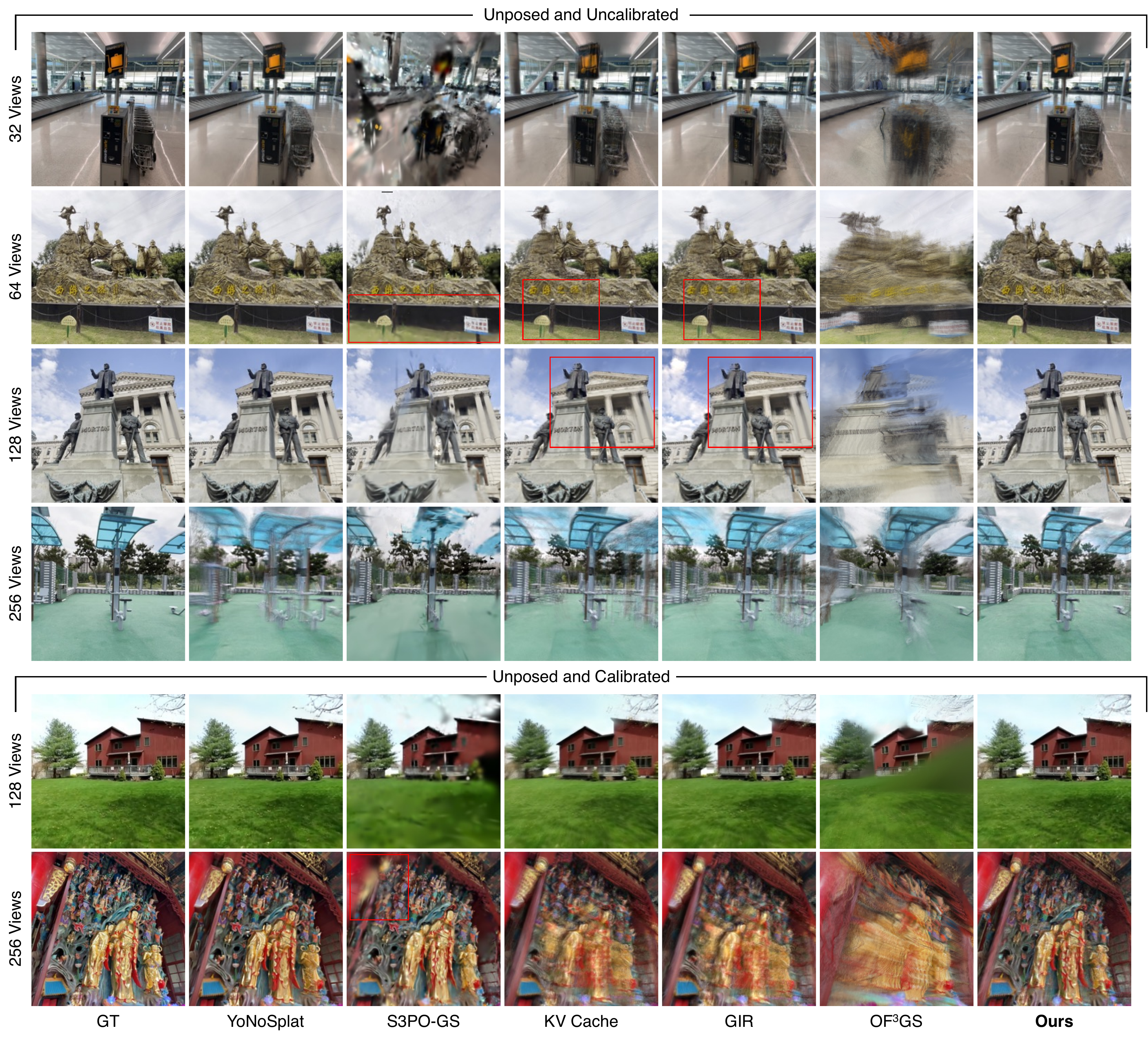}
    \caption{Novel view synthesis on DL3DV under both unposed settings. ReCoSplat consistently outperforms online baselines.}
    \label{fig:unposed}
\end{figure*}

\subsection{Experimental Setup}

\paragraph{Datasets}
We train and evaluate ReCoSplat on DL3DV \cite{ling2024dl3dv} (9894 train / 140 test) and ScanNet++ (968 train / 50 test) \cite{yeshwanthliu2023scannetpp}, sampled at a 10:1 ratio during training.
DL3DV contains diverse indoor and outdoor scenes, while ScanNet++ focuses on room-scale indoor scenes with highly irregular camera trajectories.
To assess generalization, we evaluate on RealEstate10K \cite{re10k}, ACID \cite{acid}, and ScanNet \cite{dai2017scannet}.
Due to large scene-size variance, we restrict RealEstate10K and ACID to scenes with at least 200 images (1648 for RealEstate10K, 755 for ACID).
For ScanNet, we evaluate on 8 scenes following \cite{FreeSplat} for fair comparison.
Context views are selected via farthest point sampling on ground-truth camera positions, and 8 target views are held out by sampling one frame from each of 8 uniform bins over the remaining frames.
We find evaluation metrics insensitive to the number of target views (see the supplementary material).

\paragraph{Evaluated Methods}
For novel view synthesis, we compare ReCoSplat against the online methods S3PO-GS \cite{S3PO-GS} and OF$^3$GS \cite{chen2026of3gs}.
S3PO-GS is optimization-based and requires calibrated inputs.
We do not evaluate S3PO-GS on RealEstate10K because of the per-scene optimization cost.
OF$^3$GS is feed-forward and, like ReCoSplat, supports unposed and uncalibrated inputs.
Because OF$^3$GS predicts its own intrinsics and has no input path for provided ones, we construct its calibrated variant by substituting ground-truth intrinsics into its camera decoding, which we verified leaves quality essentially unchanged.
We also treat FreeSplat \cite{FreeSplat} as an online reference, since it can be easily adapted to causal inference by restricting its cost volumes to past frames.
We evaluate the original model without this adaptation, and only on its training dataset ScanNet, as an upper bound.
The remaining online methods, StreamGS \cite{StreamGS}, SaLon3R \cite{SaLon3R}, and LongSplat \cite{huang2025longsplatonlinegeneralizable3d}, have not released code.
As a stand-in for LongSplat, we train a ReCoSplat variant that replaces the RGB+features conditioning signal in ReCo with LongSplat's Gaussian Image Representation, denoted \textbf{GIR}.
We also report a variant with the ReCo module removed entirely, denoted \textbf{KV Cache}.
As offline references, we further report YoNoSplat \cite{ye2025yonosplatneedmodelfeedforward} (fine-tuned on our training setup), AnySplat \cite{jiang2025anysplat} (unposed and uncalibrated setting), and ZPressor \cite{wang2025zpressor} (posed setting).
These methods access all views at once and are marked in gray.
For camera pose estimation, we compare with online methods spanning different memory designs: CUT3R and TTT3R (recurrent state) \cite{cut3r, chen2025ttt3r}, StreamVGGT (uncompressed KV cache) \cite{streamVGGT}, InfiniteVGGT (bounded KV cache) \cite{yuan2026infinitevggt}, and WinT3R (growing camera token pool) \cite{li2025wint3rwindowbasedstreamingreconstruction}, along with offline YoNoSplat.
In addition, we report the causally estimated poses of OF$^3$GS.

\begin{table*}[t]
\centering
\caption{Novel view synthesis on ScanNet and RealEstate10K (both out-of-distribution) and ScanNet++.
Following FreeSplat's average frame interval of 20, ScanNet inputs are from the first 640 frames; RealEstate10K and ScanNet++ input views are sampled from the entire test sequence.
For ScanNet++, we evaluate on 46/50 scenes for 256 views and 30/50 scenes for 512 views due to limited sequence length. The \highcap{best} and \medcap{second best} metrics are marked.}
\label{tab:scannetpp}
\scriptsize
\renewcommand{\arraystretch}{0.9}
\setlength{\tabcolsep}{2pt}
    \begin{tabular}{@{} lcc
        p{0.3em}
        *{3}{c} p{0.3em}
        *{3}{c} p{0.3em}
        *{3}{c} p{0.3em}
        *{3}{c} p{0.3em}
        *{3}{c} @{}}
        \toprule
        \multirow{3}{*}{Method} & \multirow{3}{*}{$p$} & \multirow{3}{*}{$k$} & & \multicolumn{3}{c}{ScanNet} & & \multicolumn{7}{c}{RealEstate10K} & & \multicolumn{7}{c}{ScanNet++} \\
        \cmidrule(lr){5-7} \cmidrule(lr){9-15} \cmidrule(lr){17-23}
        & & & & \multicolumn{3}{c}{32v} & & \multicolumn{3}{c}{64v} & & \multicolumn{3}{c}{128v} & & \multicolumn{3}{c}{256v} & & \multicolumn{3}{c}{512v} \\
        & & & & {PSNR $\uparrow$} & {SSIM $\uparrow$} & {LPIPS $\downarrow$} & & {PSNR $\uparrow$} & {SSIM $\uparrow$} & {LPIPS $\downarrow$} & & {PSNR $\uparrow$} & {SSIM $\uparrow$} & {LPIPS $\downarrow$} & & {PSNR $\uparrow$} & {SSIM $\uparrow$} & {LPIPS $\downarrow$} & & {PSNR $\uparrow$} & {SSIM $\uparrow$} & {LPIPS $\downarrow$} \\
        
        \midrule

        S3PO-GS   & & \cmark & & {15.264} & {0.487} & {0.595} & & {N/A} & {N/A} & {N/A} & & {N/A} & {N/A} & {N/A} & & {9.399} & {0.257} & {0.719} & & {9.554} & {0.272} & {0.715} \\
        OF$^3$GS  & & \cmark & & {20.449} & {0.682} & {0.369} & & {18.269} & {0.612} & {0.399} & & {17.771} & {0.597} & {0.415} & & {15.299} & {0.534} & {0.542} & & {N/A$^{\dagger}$} & {N/A$^{\dagger}$} & {N/A$^{\dagger}$} \\
        KV Cache  & & \cmark & & {25.384} & \med{0.828} & \med{0.207} & & \med{24.860} & \med{0.858} & \med{0.138} & & \med{23.640} & \med{0.820} & {0.175} & & {18.917} & {0.652} & {0.355} & & {18.155} & \med{0.625} & \med{0.405} \\
        GIR       & & \cmark & & \med{25.426} & {0.826} & {0.211} & & {24.787} & {0.856} & {0.140} & & {23.629} & \med{0.820} & \med{0.173} & & \med{19.072} & \med{0.657} & \med{0.353} & & \med{18.247} & \high{0.632} & \high{0.403} \\
        \rowcolor{grayhighlight}
        Ours      & & \cmark & & \high{25.830} & \high{0.830} & \high{0.205} & & \high{25.705} & \high{0.873} & \high{0.128} & & \high{24.713} & \high{0.844} & \high{0.155} & & \high{19.660} & \high{0.663} & \high{0.342} & & \high{18.289} & {0.611} & {0.411} \\

        \midrule

        FreeSplat & \cmark & \cmark & & {22.691} & {0.730} & {0.255} & & {N/A} & {N/A} & {N/A} & & {N/A} & {N/A} & {N/A} & & {N/A} & {N/A} & {N/A} & & {N/A} & {N/A} & {N/A} \\
        KV Cache  & \cmark & \cmark & & {23.252} & {0.761} & \med{0.245} & & \med{24.490} & \med{0.856} & \med{0.138} & & {23.179} & {0.819} & {0.175} & & {19.287} & \med{0.700} & \med{0.305} & & {18.466} & {0.680} & {0.346} \\
        GIR       & \cmark & \cmark & & \med{23.315} & \med{0.763} & {0.246} & & {24.423} & {0.854} & {0.140} & & \med{23.255} & \med{0.822} & \med{0.171} & & \med{19.331} & {0.699} & {0.308} & & \med{18.514} & \med{0.682} & \med{0.343} \\
        \rowcolor{grayhighlight}
        Ours      & \cmark & \cmark & & \high{24.073} & \high{0.789} & \high{0.223} & & \high{25.830} & \high{0.884} & \high{0.120} & & \high{24.957} & \high{0.864} & \high{0.143} & & \high{20.797} & \high{0.746} & \high{0.263} & & \high{20.308} & \high{0.734} & \high{0.287} \\
        
        \bottomrule
    \end{tabular}
\end{table*}

\paragraph{Metrics}
We evaluate novel view synthesis using PSNR, SSIM, and LPIPS \cite{zhang2018perceptual}.
In the posed setting, targets are rendered at ground-truth poses.
In the unposed setting, following prior pose-free evaluation protocols \cite{ye2025no, ye2025yonosplatneedmodelfeedforward}, target poses are initialized from scale-aligned ground-truth poses and refined for 100 iterations with the rendering loss.
For camera pose estimation, we report the area under the cumulative pose error curve (AUC) at $5^\circ$, $10^\circ$, and $20^\circ$ thresholds, where the pose error is the maximum of the rotation and translation-direction angle errors \cite{ye2025yonosplatneedmodelfeedforward, sarlin20superglue, edstedt2024roma}.

\paragraph{Implementation Details}
We train ReCoSplat at $224 \times 224$ resolution using A6000 48GB GPUs.
All training runs use 8 GPUs with a per-GPU batch size of 1.
Unless otherwise noted, inference processes streams in chunks of 8 views.
Following YoNoSplat \cite{ye2025yonosplatneedmodelfeedforward}, baselines are evaluated with inputs matching their training resolutions, and our inputs always cover a smaller or equal field of view, so resolution differences cannot favor our model.
When a baseline exceeds 48GB of GPU memory at its training resolution, we mark the cell N/A$^\dagger$ rather than substitute a smaller resolution outside its training distribution.
For evaluation, all rendered outputs are center-cropped and rescaled to $224 \times 224$.

\subsection{Main Results}
\label{sec:results}

\paragraph{Novel View Synthesis}
We evaluate ReCoSplat on novel view synthesis across three input settings: unposed and uncalibrated, unposed, and posed and calibrated.
To comprehensively assess our method, we consider multiple datasets with varying sequence lengths, scene scales, and camera trajectories. 
On DL3DV, we evaluate using 32 to 256 input views where the scene scale increases with the number of views (Table~\ref{tab:dl3dv-1}), and separately with 90 to 270 views sampled from the entire sequence (see the supplementary material).
To test robustness under highly irregular camera motion, we further evaluate on ScanNet++ with up to 512 input views (Table~\ref{tab:scannetpp}).
ScanNet++ is challenging even without the online constraint.
Even the offline YoNoSplat, fine-tuned on our training setup and hence also trained on ScanNet++, reaches only 20.989 and 20.103 PSNR at 256 and 512 views.

\begin{figure}[t]
    \centering
    \includegraphics[width=\linewidth]{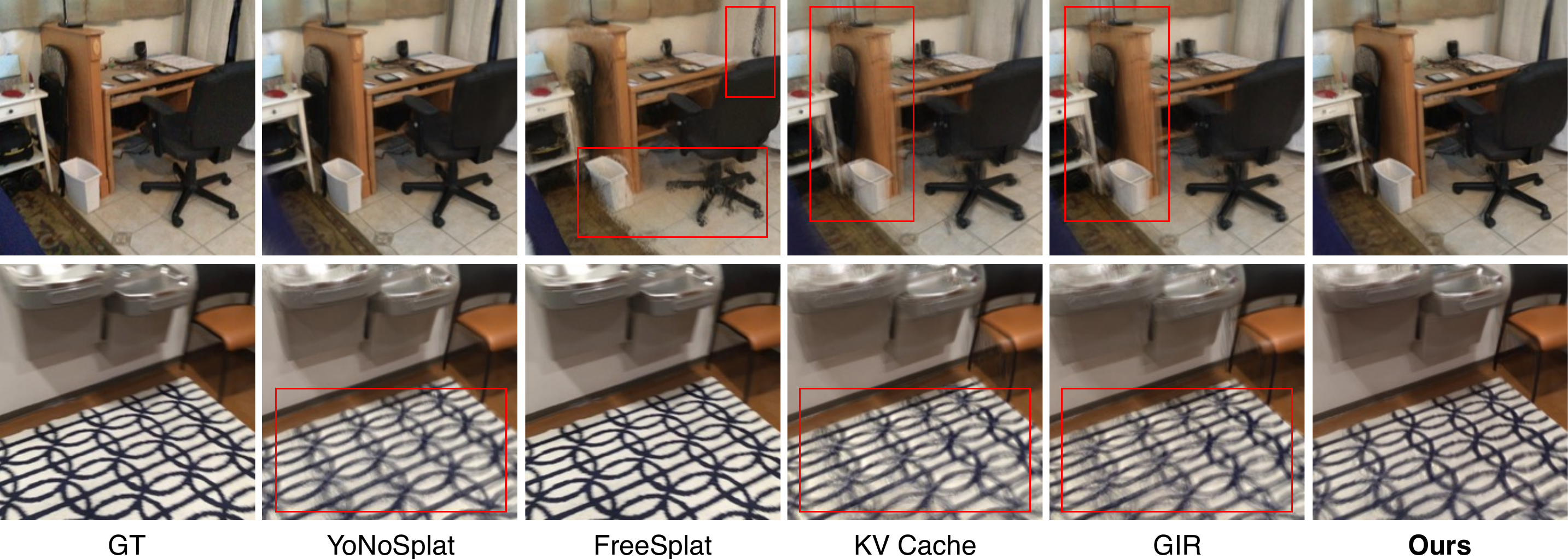}
    \caption{Novel view synthesis on ScanNet (out-of-distribution) in the posed and calibrated setting with 32 input views.}
    \label{fig:freesplat}
\end{figure}

Across all settings, ReCoSplat consistently outperforms online baselines.
\textbf{In particular, the substantial improvements in unposed settings demonstrate that our method effectively adapts to predicted assembly poses at inference time.}
Qualitatively, these improvements are reflected in Figure~\ref{fig:unposed}, where ReCoSplat produces sharper geometry and fewer artifacts than competing online methods.
On the out-of-distribution dataset ScanNet, ReCoSplat likewise produces cleaner reconstructions than FreeSplat (Figure~\ref{fig:freesplat}).
Although S3PO-GS requires known camera intrinsics, we include it in the uncalibrated rows of Figure \ref{fig:unposed} for visual comparison.
While our unposed metrics trail the offline YoNoSplat, which has access to the entire image sequence upfront, ReCoSplat delivers highly competitive performance under the strictly harder constraints of online reconstruction.

\begin{figure}[t]
    \centering
    \includegraphics[width=\linewidth,trim=0 18 0 0,clip]{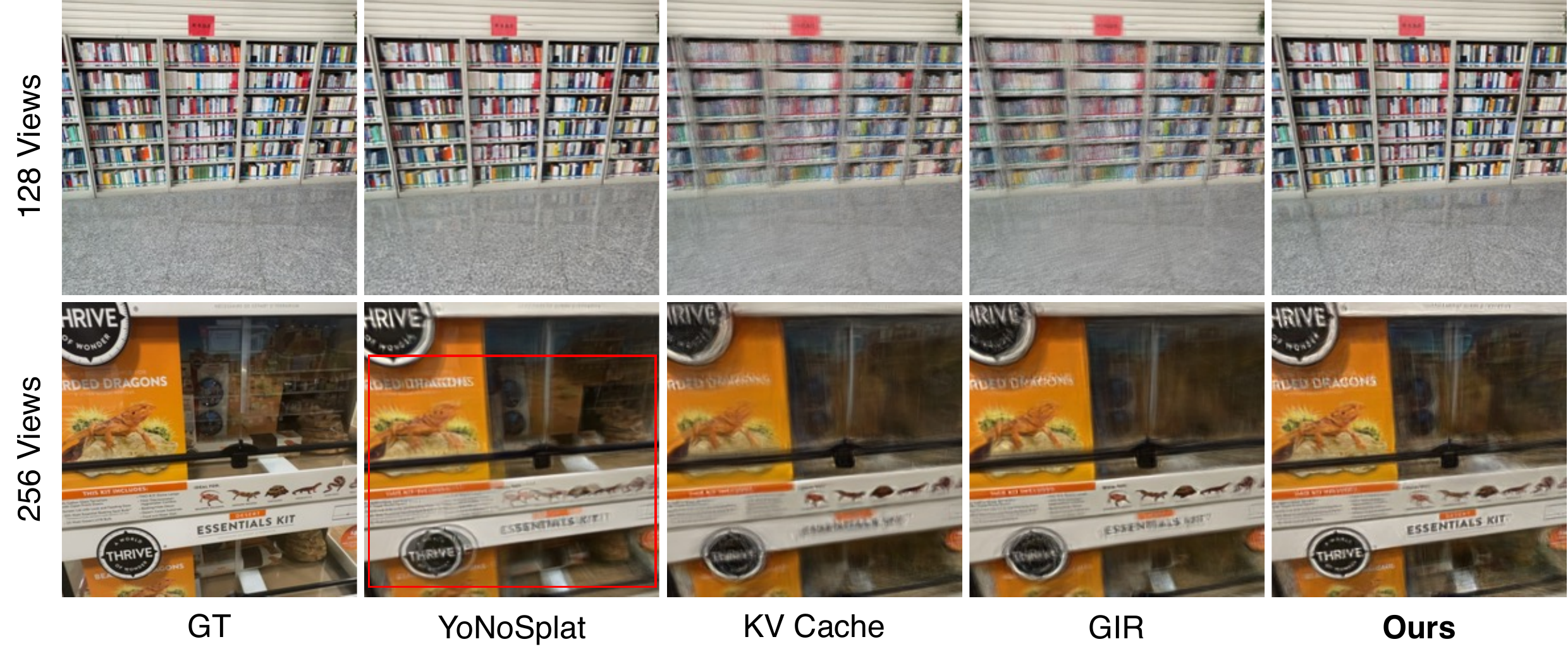}
    \caption{Novel view synthesis on DL3DV in the posed and calibrated setting with 128 and 256 input views. Without pose errors, the ReCo module corrects local Gaussian mispredictions, outperforming online baselines and surpassing YoNoSplat in PSNR.}
    \label{fig:fully-posed}
\end{figure}

In the unposed setting, online methods encounter three coupled error sources: pose distribution mismatch, local Gaussian prediction error, and pose estimation error.
\textbf{Our Render-and-Compare module targets the first two, and it is the only conditioning mechanism in our ablations that remains beneficial under predicted poses (Section~\ref{sec:ablation}).}
The effectiveness of ReCo is not limited to settings with pose discrepancies.
In fact, we find that ReCo delivers significant improvements in the posed and calibrated setting, where we approach YoNoSplat's performance and even surpass it in PSNR (Table~\ref{tab:dl3dv-1}, Figure~\ref{fig:fully-posed}).
We attribute this gain to the lack of pose prediction error in the fully posed setting.
Our posed results provide strong evidence that unposed performance will improve further as online pose estimation advances.

\begin{table*}[t]
\centering
\caption{Camera pose estimation results on ACID, RealEstate10K, and DL3DV. We report the AUC of the angular pose error curve at $5^\circ$, $10^\circ$, and $20^\circ$ thresholds for different numbers of input views. The \highcap{best} and \medcap{second best} metrics are marked, excluding \textcolor{gray}{offline} methods.}
\label{tab:pose-estimation}
\scriptsize
\renewcommand{\arraystretch}{0.9}
\setlength{\tabcolsep}{2pt}
    \begin{tabular}{@{} l
        p{0.3em}
        *{3}{c} p{0.3em}
        *{3}{c} p{0.3em}
        *{3}{c} p{0.3em}
        *{3}{c} p{0.3em}
        *{3}{c} p{0.3em}
        *{3}{c} @{}}
        \toprule
        \multirow{3}{*}{Method} & & \multicolumn{7}{c}{ACID} & & \multicolumn{7}{c}{RealEstate10K} & & \multicolumn{7}{c}{DL3DV} \\
        \cmidrule(lr){3-9} \cmidrule(lr){11-17} \cmidrule(lr){19-25}
        & & \multicolumn{3}{c}{32v} & & \multicolumn{3}{c}{64v} & & \multicolumn{3}{c}{64v} & & \multicolumn{3}{c}{128v} & & \multicolumn{3}{c}{128v} & & \multicolumn{3}{c}{256v} \\
        & & {$5^\circ \uparrow$} & {$10^\circ \uparrow$} & {$20^\circ \uparrow$} & & {$5^\circ \uparrow$} & {$10^\circ \uparrow$} & {$20^\circ \uparrow$} & & {$5^\circ \uparrow$} & {$10^\circ \uparrow$} & {$20^\circ \uparrow$} & & {$5^\circ \uparrow$} & {$10^\circ \uparrow$} & {$20^\circ \uparrow$} & & {$5^\circ \uparrow$} & {$10^\circ \uparrow$} & {$20^\circ \uparrow$} & & {$5^\circ \uparrow$} & {$10^\circ \uparrow$} & {$20^\circ \uparrow$} \\
        
        \midrule
        
        \textcolor{gray}{YoNoSplat}  & & {\textcolor{gray}{0.493}} & {\textcolor{gray}{0.670}} & {\textcolor{gray}{0.787}} & & {\textcolor{gray}{0.475}} & {\textcolor{gray}{0.661}} & {\textcolor{gray}{0.783}} & & {\textcolor{gray}{0.701}} & {\textcolor{gray}{0.846}} & {\textcolor{gray}{0.921}} & & {\textcolor{gray}{0.685}} & {\textcolor{gray}{0.834}} & {\textcolor{gray}{0.911}} & & {\textcolor{gray}{0.741}} & {\textcolor{gray}{0.870}} & {\textcolor{gray}{0.935}} & & {\textcolor{gray}{0.738}} & {\textcolor{gray}{0.869}} & {\textcolor{gray}{0.935}} \\
        CUT3R      & & {0.163} & {0.379} & {0.587} & & {0.105} & {0.312} & {0.528} & & {0.282} & {0.547} & {0.754} & & {0.140} & {0.411} & {0.666} & & {0.106} & {0.398} & {0.670} & & {0.005} & {0.024} & {0.244} \\
        TTT3R      & & \med{0.193} & \med{0.409} & \med{0.614} & & \med{0.159} & \med{0.376} & \med{0.588} & & \med{0.338} & \med{0.591} & \med{0.776} & & \med{0.270} & \med{0.539} & \med{0.748} & & {0.524} & {0.754} & {0.877} & & {0.263} & {0.579} & {0.785} \\
        StreamVGGT & & {0.028} & {0.153} & {0.382} & & {0.018} & {0.124} & {0.348} & & {0.021} & {0.164} & {0.428} & & {0.014} & {0.133} & {0.393} & & {0.012} & {0.146} & {0.427} & & {N/A$^{\dagger}$} & {N/A$^{\dagger}$} & {N/A$^{\dagger}$} \\
        InfiniteVGGT & & {0.029} & {0.151} & {0.381} & & {0.017} & {0.126} & {0.351} & & {0.023} & {0.163} & {0.426} & & {0.013} & {0.126} & {0.383} & & {0.011} & {0.140} & {0.426} & & {0.000} & {0.055} & {0.295} \\
        WinT3R     & & {0.042} & {0.177} & {0.381} & & {0.016} & {0.094} & {0.252} & & {0.186} & {0.465} & {0.700} & & {0.077} & {0.294} & {0.562} & & {0.100} & {0.372} & {0.648} & & {0.011} & {0.117} & {0.389} \\
        OF$^3$GS   & & {0.113} & {0.309} & {0.555} & & {0.089} & {0.278} & {0.532} & & {0.201} & {0.476} & {0.707} & & {0.179} & {0.456} & {0.695} & & \med{0.554} & \med{0.768} & \med{0.884} & & \med{0.505} & \med{0.746} & \med{0.873} \\
        \rowcolor{grayhighlight}
        Ours       & & \high{0.444} & \high{0.635} & \high{0.764} & & \high{0.406} & \high{0.607} & \high{0.749} & & \high{0.689} & \high{0.839} & \high{0.917} & & \high{0.666} & \high{0.823} & \high{0.905} & & \high{0.715} & \high{0.857} & \high{0.928} & & \high{0.709} & \high{0.854} & \high{0.927} \\
        
        \bottomrule
    \end{tabular}
\end{table*}

\paragraph{Camera Pose Estimation}
Table~\ref{tab:pose-estimation} reports camera pose estimation results on ACID, RealEstate10K, and DL3DV.
Among the three evaluation datasets, ReCoSplat and YoNoSplat are trained only on DL3DV, whereas CUT3R, TTT3R, and OF$^3$GS are trained on DL3DV and RealEstate10K but not ACID.
In contrast, StreamVGGT, InfiniteVGGT, and WinT3R are not trained on any of the three.
ReCoSplat consistently outperforms all online baselines across the three evaluated datasets.
In particular, on RealEstate10K, we achieve substantially higher accuracy than CUT3R and TTT3R despite these two methods being trained on RealEstate10K.
On ACID, which is unseen during training for all models, our method also achieves the strongest performance among online approaches, demonstrating strong generalization to new scene distributions.
While YoNoSplat achieves higher pose accuracy, it is an offline method with access to the entire sequence.

\begin{figure}[t]
    \centering
    \includegraphics[width=\linewidth,trim=0 6.4 0 6.3,clip]{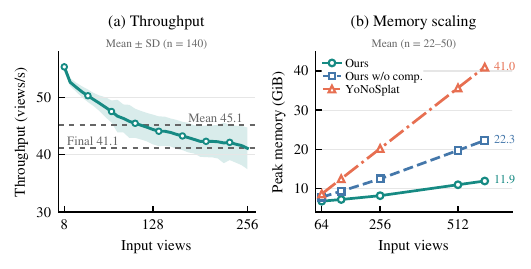}
    \caption{Runtime and memory profiling at $224 \times 224$ on an RTX 6000 Ada with chunk size 8. (a) Input throughput as the stream grows. (b) Peak memory versus the number of input views.}
    \label{fig:memory_scaling}
\end{figure}

\paragraph{Efficiency}
ReCoSplat achieves anytime rendering by accumulating an explicit set of 3D Gaussians, decoding only the incoming chunk at each update.
This keeps the update cost bounded.
Over 256-view DL3DV streams with a chunk size of 8, ReCoSplat sustains an average input throughput of 45.1 FPS, decaying only mildly to 41.1 FPS at the end of the stream (Figure~\ref{fig:memory_scaling}a).
Peak memory stays below 12 GiB on average even at 600 input views, versus 22.3 GiB without KV-cache compression and 41.0 GiB for the offline YoNoSplat (Figure~\ref{fig:memory_scaling}b).
For tttLRM to keep its scene renderable, it must instead re-decode Gaussians from all past views any time its state updates (Section~\ref{sec:related_work}).
When both methods run at $224\times224$ resolution and tttLRM's chunk size of 4, a constantly re-decoding tttLRM reaches an input throughput of only 1.4 FPS versus ReCoSplat's 32.1 FPS on an RTX 6000 Ada, a $22\times$ gap that widens with input length.

\subsection{Ablation Study}
\label{sec:ablation}

\begin{table}[t]
    \centering
    \caption{Training assembly pose and Gaussian head conditioning ablations on DL3DV with 32 input views.}
    \label{tab:ablation_pose_conditioning}
    \renewcommand{\arraystretch}{0.9}
    \setlength{\tabcolsep}{2pt}
    \resizebox{\linewidth}{!}{
        \begin{tabular}{@{}llccc p{0.3em} ccc@{}}
            \toprule
            \multirow{2}{*}{\shortstack{Assembly pose\\(training)}}
                & \multirow{2}{*}{Conditioning}
                & \multicolumn{3}{c}{Unposed ($k$)}
                & & \multicolumn{3}{c}{Posed ($p,k$)} \\
            \cmidrule(lr){3-5} \cmidrule(lr){7-9}
                & & PSNR $\uparrow$ & SSIM $\uparrow$ & LPIPS $\downarrow$
                & & PSNR $\uparrow$ & SSIM $\uparrow$ & LPIPS $\downarrow$ \\
            \midrule
            GT        & None                  & 21.346 & 0.680 & 0.224 & & 21.866 & 0.729 & 0.196 \\
            Predicted & None                  & 20.781 & 0.641 & 0.238 & & 20.836 & 0.656 & 0.229 \\
            Mixed     & None                  & 21.316 & 0.681 & 0.223 & & 21.915 & 0.736 & 0.192 \\
            GT        & Pl\"ucker rays        & 21.166 & 0.676 & 0.228 & & 22.067 & 0.744 & 0.189 \\
            GT        & ReCo, RGB only        & 21.455 & 0.685 & 0.223 & & 22.548 & 0.759 & 0.181 \\
            GT        & ReCo, RGB + features  & 21.769 & 0.699 & 0.212 & & 22.610 & 0.759 & 0.178 \\
            \bottomrule
        \end{tabular}
    }
\end{table}

\begin{table}[t]
    \centering
    \caption{KV-cache compression ablation on unposed ($k$) DL3DV with chunk size 8.
    Both checkpoints are fully trained Stage-3 models under their respective compression policies.}
    \label{tab:ablation_kv_cache}
    \renewcommand{\arraystretch}{0.9}
    \setlength{\tabcolsep}{2pt}
    \resizebox{\linewidth}{!}{
    \begin{tabular}{@{}l ccc p{0.3em} ccc p{0.3em} ccc p{0.3em} ccc@{}}
        \toprule
        \multirow{2}{*}{KV cache} & \multicolumn{3}{c}{32v} & & \multicolumn{3}{c}{64v} & & \multicolumn{3}{c}{128v} & & \multicolumn{3}{c}{256v} \\
        \cmidrule(lr){2-4} \cmidrule(lr){6-8} \cmidrule(lr){10-12} \cmidrule(lr){14-16}
        & {\scriptsize PSNR $\uparrow$} & {\scriptsize SSIM $\uparrow$} & {\scriptsize LPIPS $\downarrow$} & & {\scriptsize PSNR $\uparrow$} & {\scriptsize SSIM $\uparrow$} & {\scriptsize LPIPS $\downarrow$} & & {\scriptsize PSNR $\uparrow$} & {\scriptsize SSIM $\uparrow$} & {\scriptsize LPIPS $\downarrow$} & & {\scriptsize PSNR $\uparrow$} & {\scriptsize SSIM $\uparrow$} & {\scriptsize LPIPS $\downarrow$} \\
        \midrule
        Full cache & 22.575 & 0.742 & 0.182 & & 22.233 & 0.724 & 0.198 & & 21.574 & 0.693 & 0.224 & & 20.222 & 0.624 & 0.279 \\
        Compressed & 22.417 & 0.734 & 0.188 & & 22.068 & 0.713 & 0.206 & & 21.576 & 0.690 & 0.227 & & 20.430 & 0.633 & 0.275 \\
        \bottomrule
    \end{tabular}
    }
\end{table}

We ablate ReCoSplat's key design choices on DL3DV.
Table~\ref{tab:ablation_pose_conditioning} ablates the training assembly pose and the Gaussian head conditioning signal.
Models in this table are trained with up to 32 views to reduce training time.
Table~\ref{tab:ablation_kv_cache} ablates KV-cache compression.

\paragraph{Training Assembly Pose (Table~\ref{tab:ablation_pose_conditioning})}
Training with predicted assembly poses is significantly worse than with ground-truth poses.
This is likely because the former couples local Gaussian prediction with pose estimation.
Meanwhile, mixing ground-truth and predicted assembly poses, following \cite{ye2025yonosplatneedmodelfeedforward}, does not improve over using ground truth poses alone, likely because pose estimation in the online setting is harder than offline.
The pose-distribution mismatch therefore cannot be solved by mixing assembly poses during training.

\paragraph{Conditioning Signal (Table~\ref{tab:ablation_pose_conditioning})}
To isolate the value of conditioning on renders of the accumulated scene, we train a model that instead conditions the Gaussian head on Pl\"ucker raymaps of the assembly poses, which encode per-pixel camera rays but no scene content.
This helps the posed setting but equally hurts the unposed one, whereas ReCo improves both.
We also observe that adding learned feature channels to ReCo is helpful in the unposed setting.

\paragraph{KV-Cache Compression (Table~\ref{tab:ablation_kv_cache})}
We find our KV-cache compression strategy to have a small effect on reconstruction quality at smaller view counts.
On the other hand, we find compression to be beneficial at higher view counts while reducing memory consumption at inference (Figure~\ref{fig:memory_scaling}b).
While we find our trainable register tokens, which mark retained views in the KV cache, to have a neutral effect on DL3DV, we find them consistently beneficial on ScanNet++'s irregular pose trajectories (please see the supplementary material).

\paragraph{Chunk-Size Curriculum}
We find that our chunk-size curriculum makes inference at variable chunk sizes possible with minimal quality variance, at no loss compared to single-chunk-size training when evaluated at that chunk size (please see the supplementary material).

\section{Conclusion}
In this work, we present ReCoSplat, an online feed-forward Gaussian Splatting method.
We introduce a Render-and-Compare module to address the mismatch between the ground-truth assembly poses used in training and the predicted ones used at inference, improving reconstruction in both posed and unposed settings.
To support long input sequences, we further propose a KV-cache compression strategy that greatly reduces memory usage while maintaining reconstruction quality.
Experiments across five datasets show that ReCoSplat achieves real-time, state-of-the-art online novel view synthesis and camera pose estimation.

\paragraph{Acknowledgements}
We thank Sifei Liu for extensive discussions and insightful suggestions that helped improve this work.
We also thank Shuo Chen and Kuan-Chih Huang for helpful discussions.

{
    \small
    \pdfbookmark[1]{References}{references}
    \bibliographystyle{ieeenat_fullname}
    \bibliography{main}
}

\clearpage
\pdfbookmark[0]{Supplementary Material}{supplementary}
\maketitlesupplementary

\appendix

\section{Appendix}

Unless otherwise noted, all ablation experiments are run on A6000 48GB, A100 40GB, or H100 80GB GPUs, depending on availability.

\subsection{Limitations}

While ReCoSplat's Render-and-Compare module helps bridge the pose distribution mismatch, reconstruction quality in unposed settings cannot be fully independent of pose estimation accuracy.
Large pose errors can still propagate to Gaussian assembly and affect rendering fidelity.
Advances in online pose estimation are therefore likely to further strengthen online reconstruction performance.

\subsection{Additional Ablation Studies}

\begin{figure*}[tbp]
    \centering
    \includegraphics[width=\linewidth]{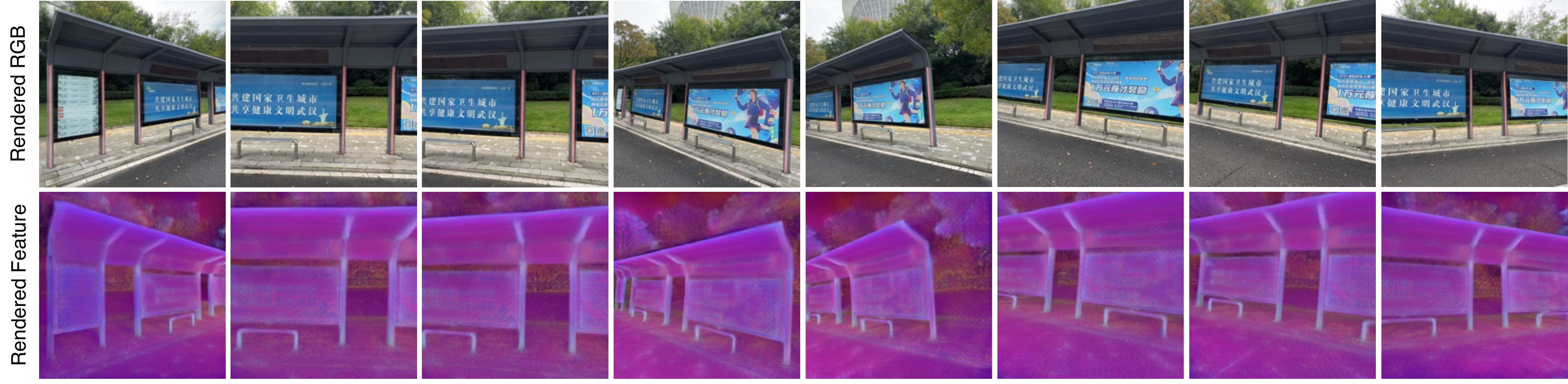}
    \caption{Visualization of the additional Gaussian feature channels rendered by our model, reduced to three dimensions via PCA projection.}
    \label{fig:supp_feature_vis}
\end{figure*}

\paragraph{Gaussian Features}
The Render-and-Compare module renders the accumulated Gaussian scene from the assembly pose of each incoming view and uses the result to condition the prediction of new local Gaussians.
In addition to RGB, each Gaussian carries nine learned feature dimensions, which are rendered from the accumulated scene along with its color.
These feature maps give the model a richer description of the existing reconstruction than RGB alone.
Figure~\ref{fig:supp_feature_vis} visualizes the rendered features after projecting them to three dimensions with PCA.

\paragraph{Trainable Register Tokens}
The decoder contains 18 global-attention layers.
Our KV-cache compression strategy disables caching in the first 10 global-attention layers, where attention is restricted to the current chunk.
For each of the last 8 global-attention layers, the cache retains the keys and values of all views in the first chunk and only the final view of every subsequent chunk.
This preserves the initial scene context and a sparse history of later observations while substantially reducing the cache size.

In addition to the decoder's standard register tokens, we learn two retention-specific embeddings, one shared by the retained views from the first chunk and one shared by the retained final views of later chunks.
These embeddings mark how the tokens in the compressed history were retained.
Table~\ref{tab:supp_register_token} isolates this component.
The ablated model uses the same KV-cache compression strategy and standard register tokens, but does not use the retention-specific embeddings.
The tokens have little effect on DL3DV~\cite{ling2024dl3dv}, whereas they consistently improve all reported metrics on ScanNet++~\cite{yeshwanthliu2023scannetpp}.
This dataset-dependent difference suggests that explicitly marking retained views is more useful for streams with irregular camera trajectories.

\begin{table*}[tbp]
\centering
\caption{Effect of the retention-specific register-token embeddings. The ablated model uses the same KV-cache compression strategy and standard register tokens, but does not use the specialized embeddings that mark retained views. A check mark under $p$ or $k$ indicates that camera extrinsics or intrinsics are provided, respectively. DL3DV results average 140 scenes. ScanNet++ results use 50 scenes at 128 views, 46 scenes at 256 views, and 30 scenes at 512 views because some sequences are too short for the larger view counts. The \highcap{best} metrics are marked.}
\label{tab:supp_register_token}
\footnotesize
\renewcommand{\arraystretch}{1.15}
\setlength{\tabcolsep}{2pt}

\textbf{(a) DL3DV}\\[0.35em]
\begin{tabular}{@{} lcc
    p{0.5em}
    *{3}{c} p{0.5em}
    *{3}{c} p{0.5em}
    *{3}{c} @{}}
    \toprule
    \multirow{2}{*}{Method} & \multirow{2}{*}{$p$} & \multirow{2}{*}{$k$} & & \multicolumn{3}{c}{90v} & & \multicolumn{3}{c}{180v} & & \multicolumn{3}{c}{270v} \\
    & & & & {PSNR $\uparrow$} & {SSIM $\uparrow$} & {LPIPS $\downarrow$} & & {PSNR $\uparrow$} & {SSIM $\uparrow$} & {LPIPS $\downarrow$} & & {PSNR $\uparrow$} & {SSIM $\uparrow$} & {LPIPS $\downarrow$} \\
    \midrule
    Ours w/o Retained-View Marking & & \cmark & & {21.175} & {0.669} & {0.238} & & {20.820} & {0.649} & {0.261} & & \high{20.365} & \high{0.626} & {0.284} \\
    Ours & & \cmark & & \high{21.206} & \high{0.670} & \high{0.237} & & \high{20.823} & \high{0.651} & \high{0.259} & & {20.325} & {0.625} & \high{0.282} \\
    \bottomrule
\end{tabular}

\vspace{0.8em}
\textbf{(b) ScanNet++}\\[0.35em]
\begin{tabular}{@{} lcc
    p{0.5em}
    *{3}{c} p{0.5em}
    *{3}{c} p{0.5em}
    *{3}{c} @{}}
    \toprule
    \multirow{2}{*}{Method} & \multirow{2}{*}{$p$} & \multirow{2}{*}{$k$} & & \multicolumn{3}{c}{128v} & & \multicolumn{3}{c}{256v} & & \multicolumn{3}{c}{512v} \\
    & & & & {PSNR $\uparrow$} & {SSIM $\uparrow$} & {LPIPS $\downarrow$} & & {PSNR $\uparrow$} & {SSIM $\uparrow$} & {LPIPS $\downarrow$} & & {PSNR $\uparrow$} & {SSIM $\uparrow$} & {LPIPS $\downarrow$} \\
    \midrule
    Ours w/o Retained-View Marking & & \cmark & & {19.635} & {0.675} & {0.337} & & {19.492} & {0.654} & {0.351} & & {17.791} & {0.594} & {0.438} \\
    Ours & & \cmark & & \high{19.840} & \high{0.683} & \high{0.331} & & \high{19.660} & \high{0.663} & \high{0.342} & & \high{18.289} & \high{0.611} & \high{0.411} \\
    \bottomrule
\end{tabular}
\end{table*}

\paragraph{Chunk-Size Curriculum Learning}
To ablate chunk-size curriculum learning, we train a baseline using only chunk size 8.
Our curriculum instead anneals the minimum sampled chunk size from 8 to 4 during Stage~2.
Table~\ref{tab:supp_chunk_curriculum} shows that the curriculum-trained model matches the baseline when both are evaluated at chunk size 8.
When later chunks are reduced to four views, unposed PSNR decreases by only $0.13$~dB for the curriculum-trained model, compared with $0.73$~dB for the baseline.
The curriculum therefore preserves quality at the original chunk size while enabling inference with different chunk sizes.

\begin{table*}[tbp]
    \centering
    \caption{Chunk-size curriculum learning (CL) ablation on DL3DV with 128 input views. Two Stage-2 checkpoints, trained with and without the curriculum, are each evaluated using either 8 views in every chunk or 8 views in the first chunk followed by 4 views in every later chunk. The Stage-3 model is included only for reference.}
    \label{tab:supp_chunk_curriculum}
    \footnotesize
    \renewcommand{\arraystretch}{1.15}
    \setlength{\tabcolsep}{2pt}
    \begin{tabular}{@{} lc
        p{0.5em}
        *{3}{c} p{0.5em}
        *{3}{c} @{}}
        \toprule
        \multirow{2}{*}{Method} & \multirow{2}{*}{Stage} & &
            \multicolumn{3}{c}{Unposed ($k$)} & & \multicolumn{3}{c}{Posed ($p,k$)} \\
        \cmidrule(lr){4-6} \cmidrule(lr){8-10}
            & & & PSNR $\uparrow$ & SSIM $\uparrow$ & LPIPS $\downarrow$ & & PSNR $\uparrow$ & SSIM $\uparrow$ & LPIPS $\downarrow$ \\
        \midrule
        Ours                                      & 2 & & 21.527 & 0.689 & 0.227 & & 22.908 & 0.780 & 0.174 \\
        Ours w/o CL                               & 2 & & 21.510 & 0.688 & 0.227 & & 22.884 & 0.779 & 0.175 \\
        Ours @ Chunk Size $(8,4,4,\ldots)$        & 2 & & 21.397 & 0.680 & 0.237 & & 22.816 & 0.775 & 0.181 \\
        Ours w/o CL @ Chunk Size $(8,4,4,\ldots)$ & 2 & & 20.777 & 0.652 & 0.265 & & 21.853 & 0.740 & 0.215 \\
        \midrule
        Ours                                      & 3 & & 21.576 & 0.690 & 0.227 & & 22.852 & 0.777 & 0.176 \\
        \bottomrule
    \end{tabular}
\end{table*}

\paragraph{Scaling to $518\!\times\!518$}
We briefly fine-tune ReCoSplat at $518\!\times\!518$ to measure its high-resolution inference efficiency.
Due to compute constraints, this checkpoint is not fully converged and is used only for throughput and memory profiling, not reconstruction-quality comparisons.
Across 47 256-view ScanNet++ streams with chunk size 8, ReCoSplat processes inputs at 14.4 FPS on average and 13.2 FPS for the final chunk on a single H100 80GB GPU.
Its mean peak allocated GPU memory is 23.5 GiB.

\subsection{Additional Novel View Synthesis Results}

\paragraph{DL3DV Whole-Sequence Results}
In Table~1, the 32-, 64-, 128-, and 256-view inputs are sampled from the first 180, 240, 300, and up to 360 frames of each scene, respectively.
Thus, both the number of input views and the portion of the sequence covered by those views increase across the table.
Table~\ref{tab:supp_dl3dv_whole} instead distributes 90, 180, or 270 input views across the entire sequence, so every column covers the same full sequence using a different number of observations.
OF$^3$GS~\cite{chen2026of3gs} uses $518\!\times\!518$ inputs at 90 and 180 views and $392\!\times\!518$ inputs at 270 views.
Both input shapes are sampled during OF$^3$GS training.
ReCoSplat outperforms all reported online baselines in both the unposed and posed settings at every view count.

\begin{table*}[tbp]
\centering
\caption{Novel view synthesis on all 140 DL3DV test scenes with 90, 180, and 270 input views sampled from the entire sequence. A check mark under $p$ or $k$ indicates that camera extrinsics or intrinsics are provided, respectively. The \highcap{best} and \medcap{second best} reported metrics are marked.}
\label{tab:supp_dl3dv_whole}
\footnotesize
\renewcommand{\arraystretch}{1.15}
\setlength{\tabcolsep}{2pt}
\begin{tabular}{@{} lcc
    p{0.5em}
    *{3}{c} p{0.5em}
    *{3}{c} p{0.5em}
    *{3}{c} @{}}
    \toprule
    \multirow{2}{*}{Method} & \multirow{2}{*}{$p$} & \multirow{2}{*}{$k$} & & \multicolumn{3}{c}{90v} & & \multicolumn{3}{c}{180v} & & \multicolumn{3}{c}{270v} \\
    & & & & {PSNR $\uparrow$} & {SSIM $\uparrow$} & {LPIPS $\downarrow$} & & {PSNR $\uparrow$} & {SSIM $\uparrow$} & {LPIPS $\downarrow$} & & {PSNR $\uparrow$} & {SSIM $\uparrow$} & {LPIPS $\downarrow$} \\
    \midrule
    S3PO-GS~\cite{S3PO-GS} & & \cmark & & {15.358} & {0.380} & {0.586} & & {16.104} & {0.415} & {0.561} & & {15.700} & {0.393} & {0.572} \\
    OF$^3$GS & & \cmark & & {16.410} & {0.431} & {0.515} & & {15.976} & {0.417} & {0.537} & & {15.642} & {0.403} & {0.550} \\
    KV Cache & & \cmark & & {20.703} & {0.654} & {0.248} & & {20.318} & {0.635} & {0.269} & & {19.922} & \med{0.614} & \med{0.293} \\
    GIR      & & \cmark & & \med{20.886} & \med{0.662} & \med{0.244} & & \med{20.461} & \med{0.640} & \med{0.268} & & \med{19.949} & {0.613} & {0.294} \\
    \rowcolor{grayhighlight}
    Ours     & & \cmark & & \high{21.206} & \high{0.670} & \high{0.237} & & \high{20.823} & \high{0.651} & \high{0.259} & & \high{20.325} & \high{0.625} & \high{0.282} \\
    \midrule
    KV Cache & \cmark & \cmark & & {21.478} & {0.729} & \med{0.206} & & {21.104} & {0.716} & \med{0.219} & & {20.558} & \med{0.694} & \med{0.240} \\
    GIR      & \cmark & \cmark & & \med{21.556} & \med{0.730} & \med{0.206} & & \med{21.166} & \med{0.717} & \med{0.219} & & \med{20.598} & \med{0.694} & {0.241} \\
    \rowcolor{grayhighlight}
    Ours     & \cmark & \cmark & & \high{22.408} & \high{0.758} & \high{0.188} & & \high{22.280} & \high{0.759} & \high{0.193} & & \high{21.866} & \high{0.748} & \high{0.206} \\
    \bottomrule
\end{tabular}
\end{table*}

\paragraph{Number of Target Views}
Our evaluations render 8 held-out target views per scene.
To determine whether this choice materially affects the aggregate metric, Table~\ref{tab:supp_target_count} repeats the 128-view DL3DV evaluation with 8, 16, 24, and 32 held-out targets while keeping the input context views fixed.
Each column averages the same 140 test scenes.
PSNR changes by about 0.03\,dB, indicating little sensitivity to the number of targets.
This auxiliary experiment uses a separately generated target set from Table~1, so its 8-target value is not expected to match the corresponding value in that table exactly.

\begin{table}[tbp]
    \centering
    \caption{Sensitivity to the number of held-out target views on all 140 DL3DV test scenes with 128 input views in the unposed, calibrated ($k$) setting. The context views are fixed across columns.}
    \label{tab:supp_target_count}
    \footnotesize
    \renewcommand{\arraystretch}{1.15}
    \begin{tabular}{@{} l cccc @{}}
        \toprule
        Target views & 8 & 16 & 24 & 32 \\
        \midrule
        PSNR $\uparrow$ & 21.551 & 21.547 & 21.565 & 21.577 \\
        \bottomrule
    \end{tabular}
\end{table}

\subsection{Additional Visual Comparisons}
\begin{figure*}[b]
    \centering
    \includegraphics[width=\linewidth]{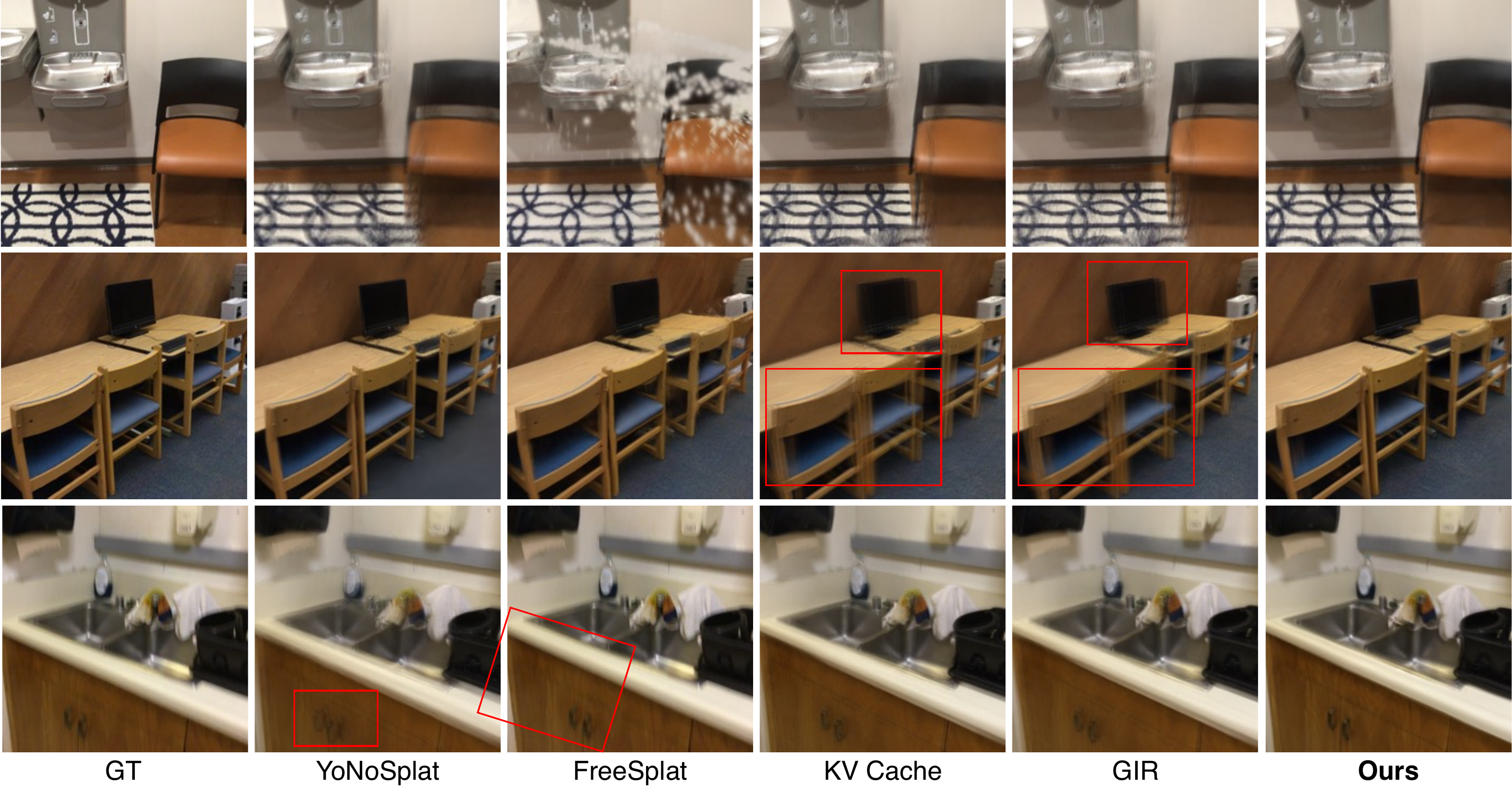}
    \caption{Novel view synthesis on the out-of-distribution ScanNet dataset with 32 input views and provided camera intrinsics and extrinsics.}
    \label{fig:supp_scannet_qual}
\end{figure*}

Figure~\ref{fig:supp_scannet_qual} compares novel view synthesis on ScanNet~\cite{dai2017scannet}, which is not used for training, using 32 posed and calibrated input views.
ReCoSplat preserves finer image structure and produces fewer visible artifacts than the online baselines and FreeSplat~\cite{FreeSplat}.

\begin{figure*}[p]
    \centering
    \includegraphics[width=\linewidth,totalheight=0.92\textheight,keepaspectratio]{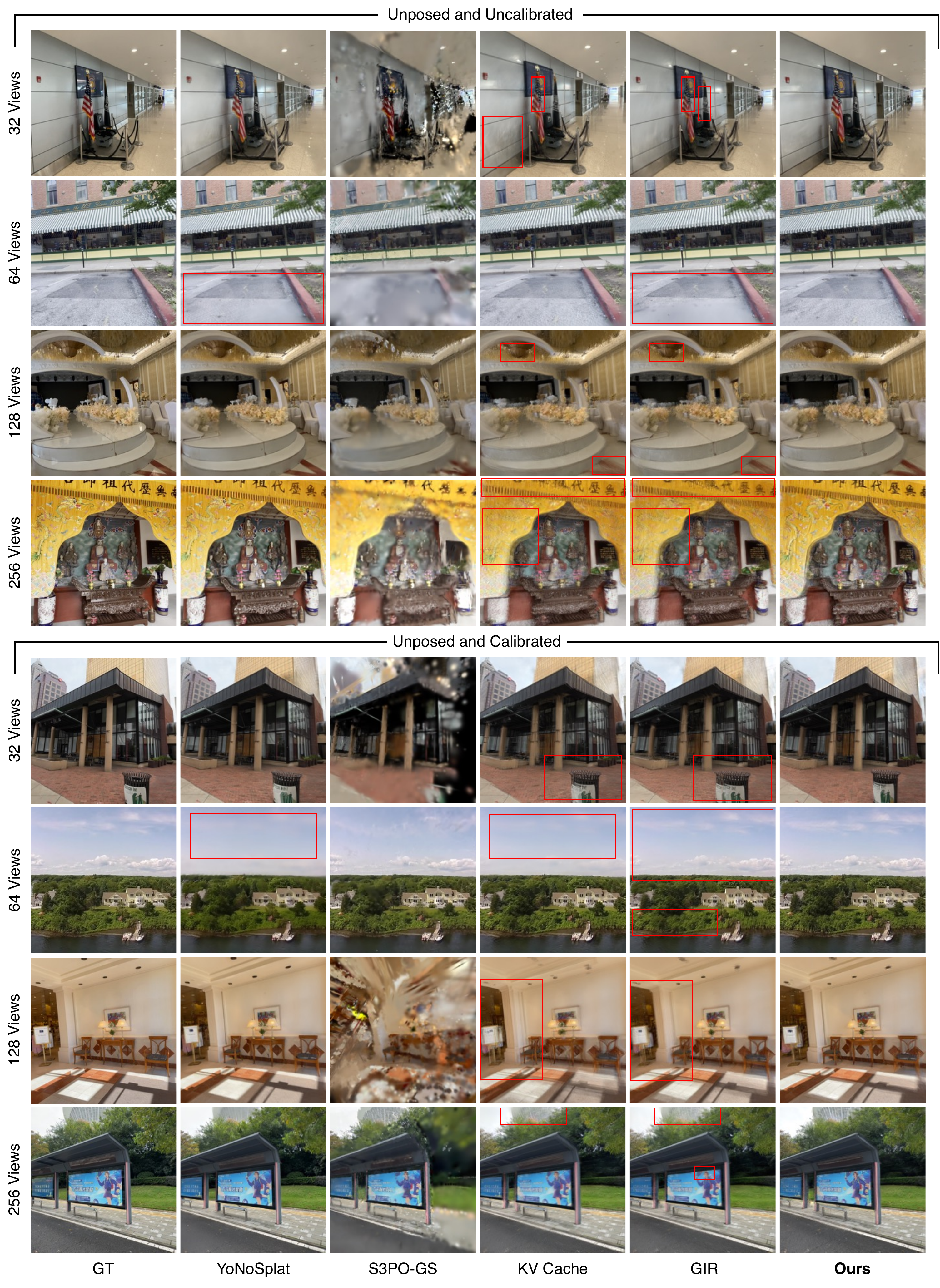}
    \caption{Novel view synthesis on DL3DV with 32 to 256 input views in the unposed and uncalibrated setting (top) and the unposed and calibrated setting (bottom). S3PO-GS normally requires calibrated inputs; we nevertheless include its output in the uncalibrated comparison for reference.}
    \label{fig:supp_dl3dv_qual}
\end{figure*}

Figure~\ref{fig:supp_dl3dv_qual} provides the corresponding comparison on DL3DV for streams ranging from 32 to 256 input views.
We show both unposed settings, one without camera intrinsics and one with ground-truth intrinsics.

\subsection{Loss Formulation}

We combine pixel MSE, LPIPS~\cite{zhang2018perceptual}, intrinsic, extrinsic, and opacity losses with weights $1.0$, $0.05$, $0.5$, $0.1$, and $0.01$, respectively.

\paragraph{Rendering Losses}
The pixel loss is MSE between rendered and ground-truth images at eight held-out target views.
The perceptual loss is VGG-based LPIPS over the same views.

\paragraph{Intrinsic Loss}
The intrinsic loss is MSE between the predicted and ground-truth horizontal and vertical focal lengths, normalized by the image dimensions.

\paragraph{Opacity Loss}
The opacity loss is an $\ell_1$ penalty on the activated Gaussian opacities.
At inference, we prune Gaussians with opacity at or below $0.005$.

\paragraph{Relative Pose Loss}
$\mathcal{L}_{\mathrm{extrinsic}}$ follows the pairwise relative-pose loss of YoNoSplat~\cite{ye2025yonosplatneedmodelfeedforward}.
For ground-truth poses $P_i=[R_i,t_i]$ and predictions $\hat P_i=[\hat R_i,\hat t_i]$, let $P_{i\leftarrow j}=P_i^{-1}P_j$ and $\hat P_{i\leftarrow j}=\hat P_i^{-1}\hat P_j$ denote the relative transformations between views $i$ and $j$.
Because translation is only determined up to scale, we multiply every predicted translation by the ratio between the maximum pairwise camera-center distances of the ground-truth and predicted trajectories.
We average the rotation and translation errors over all ordered pairs of distinct views in a sequence of length $N$.
\begin{align*}
    \mathcal{L}_{\mathrm{extrinsic}}
    &= \frac{1}{N(N-1)}\sum_{i\ne j}
    \left(\mathcal{L}_{R}(i,j)+\lambda_t\mathcal{L}_{t}(i,j)\right), \\
    \mathcal{L}_{R}(i,j)
    &= \arccos\!\left(\frac{\operatorname{tr}(R_{i\leftarrow j}^{\top}\hat R_{i\leftarrow j})-1}{2}\right), \\
    \mathcal{L}_{t}(i,j)
    &= \mathcal{H}_{\delta}(\hat t_{i\leftarrow j}-t_{i\leftarrow j}).
\end{align*}
where $\mathcal{H}_{\delta}$ is the Huber loss, $\lambda_t=100$, and $\delta=0.1$.

\end{document}